\documentclass[11pt,english]{paper}
\usepackage[T1]{fontenc}
\usepackage[latin9]{inputenc}
\usepackage{geometry}
\usepackage{babel}
\usepackage{amsthm}
\usepackage{amsmath}
\usepackage{amssymb}
\usepackage{graphicx}
\usepackage{setspace}
\usepackage[numbers]{natbib}
\usepackage[unicode=true,pdfusetitle,
 bookmarks=true,bookmarksnumbered=false,bookmarksopen=false,
 breaklinks=false,pdfborder={0 0 1},backref=false,colorlinks=false]
 {hyperref}

\makeatletter

\DeclareRobustCommand{\greektext}{%
  \fontencoding{LGR}\selectfont\def\encodingdefault{LGR}}
\DeclareRobustCommand{\textgreek}[1]{\leavevmode{\greektext #1}}
\DeclareFontEncoding{LGR}{}{}
\DeclareTextSymbol{\~}{LGR}{126}
\newcommand{\lyxmathsym}[1]{\ifmmode\begingroup\def\b@ld{bold}
  \text{\ifx\math@version\b@ld\bfseries\fi#1}\endgroup\else#1\fi}

\let\SF@@footnote\footnote
\def\footnote{\ifx\protect\@typeset@protect
    \expandafter\SF@@footnote
  \else
    \expandafter\SF@gobble@opt
  \fi
}
\expandafter\def\csname SF@gobble@opt \endcsname{\@ifnextchar[
  \SF@gobble@twobracket
  \@gobble
}
\edef\SF@gobble@opt{\noexpand\protect
  \expandafter\noexpand\csname SF@gobble@opt \endcsname}
\def\SF@gobble@twobracket[#1]#2{}

\numberwithin{equation}{section}
\numberwithin{figure}{section}

\makeatother

\begin{document}
\begin{singlespace}
\title{Between theory and practice: guidelines for an optimization scheme
with genetic algorithms - Part I: single-objective continuous global
optimization}
\end{singlespace}

\author{Loris Serafino\thanks{For correspondence: loris.serafino@kuang-chi.org} \\
Kuang-Chi Institute of Advanced Technology, Shenzhen, China}

\maketitle
\begin{abstract}
The rapid advances in the field of optimization methods in many pure
and applied science pose the difficulty of keeping track of the developments
as well as selecting an appropriate technique that best suits the
problem in-hand. From a practitioner point of view is rightful to
wander {}``which optimization method is the best for my problem?''.
Looking at the optimization process as a {}``system'' of interconnected
parts, in this paper are collected some ideas about how to tackle
an optimization problem using a class of tools from evolutionary computations
called Genetic Algorithms. Despite the number of optimization techniques
available nowadays the author of this paper thinks that Genetic Algorithms
still play a central role for their versatility, robustness, theoretical
framework and simplicity of use. The paper can be considered a {}``collection
of tips'' (from literature and personal experience) for the non-computer-scientist
that has to deal with optimization problems both in the science and
engineering practice. No original methods or algorithms are proposed.
\end{abstract}

\section{Introduction: optimization-as-a-process}

Generally in literature, with the term \textquotedbl{}optimization\textquotedbl{}
is related to (the output of) a mathematical technique or algorithm
used to identify the extreme value of an arbitrary objective function
through the manipulation of a known set of variables and subject to
a list of constraining relationships. These mathematical and algorithmic
techniques are the focus of the majority of optimization literature.
In this paper we want to look at the this issue from a more abstract
stance. In this context {}``optimization'' also has a more general
usage referring to a system of physical objects (computers, machines)
abstract ideas and people combined for discovering and verifying the
best implementable solution to high-dimensional design problems in
across a broad range of engineering and sciences application. In this
more general usage, optimization can be conceptualized as a (complex)
system composed with different sub-systems is a multiphase process,
in which applying a particular maximization algorithm or technique
is but one step. The success or the failure of the optimization process
can depend on a single step (right implementation of a algorithm e.
g. what we refer to as the {}``maximization step'') or for a not
correct interaction of the different parts. So in what follow we will refer as the technical-mathematical part of the optimization process as the:\medskip{}.

\noindent \textbf{Maximization Problem}
\footnote{Of course it is always possible to transform a minimization problem
into a maximization one trough the transformation $g(x)=-f(x)$.}. 
A maximization problem with an explicit objective can in general be expressed in the following generic form:
\[
\max_{x\in\mathcal{H}}f(x),
\]

\noindent where $x$ is a give vector in a generic multidimensional
space $\mathcal{H}$ and $f:\mathcal{H\rightarrow\mathbb{R}}$ is
a scalar function of the vector $x$ and $\mathcal{H}\subset\mathbb{R}^{n}$is
a (discrete or continuous) subset of the multidimensional real Euclidean
space. From now on we will refer to $\mathcal{H}$ as the \emph{search
space}\footnote{Even if there is no explicit mention of constrains here the formulation is nonetheless enough general since they can be incorporated through an appropriate definition of the search space $\mathcal{H}$}.\medskip{}

According to systems theorists a system is comprised of a large number
of (possibly non-linearly) interacting elements \citep{Bertalla}.
The task to decompose the optimization system into smaller, simpler
sub-elements is partly arbitrary and can produce different conceptualizations. 

The starting point to the systemic conceptualization of the optimization
process is the basic fact that optimization algorithms don't exist
in isolation but they represent one side of a coin where the other
side is represented by a problem (a function, fitness or cost) to
be optimized. In turn a fitness can be the mathematical approximation
of a physical process or system. There is no space here to fully develop
the implication of a system-theoretic approach to the optimization
process. Nonetheless this framework will stay at the background in
the consideration that will follow. Two premises are important to
be stressed at the outset. First the view of the optimization practice
as a self-adjusting process. In a given application context can be
useful to start with class of optimization algorithms and testing
them with the problem at hand. The information collected about the problem 
and the response obtained from algorithms can allow updating both
the parameters configuration of the single algorithm, the algorithm
used and the overall optimization strategy itself. The {}``optimization-as-adaptive-process''
framework means basically that maximization operations and knowledge
about a give problem must precede hand-in-hand, like two sides of
a coin, and that the information from one side can help improvements
in the other. Regarding the specif algorithm to apply today the spectrum
of methods for solving optimization problems are actually vast. Just
to name a few: genetic algorithms, simulated annealing, tabu search,
particle swarm, ant colony optimization, cross-entropy, etc. From
a practitioner point of view is rightful to wander {}``which optimization
method is the best for our problem?''. According to the most common
understanding of the so called \emph{No Free Lunch Theorems} \citep{wolma}
there is no optimization method superior to others for all possible
optimization problems. Moreover, an algorithm that performs well on
one class of problems must perform \emph{worse than random search}
on all remaining problems. Running an algorithm on a small number
of problems with a small range of parameter settings may not be a
good predictor of that algorithm's performance on other problems,
especially problems of a different type. It follows also that \textbf{running
after the last algorithm published in the literature claiming to be
able to outperform in this or that class of test problems is not necessarily
a clever approach}. As a starting overall {}``optimization philosophy''
the following steps can be followed \citep{yang}:
\begin{itemize}
\item The first approach essentially consists of using calculus tools on
the target problem. If the function is simple, use the stationary conditions (first derivatives must be zero) and extreme points (boundaries) to find the optimal solution(s). If this is not possible then some well established conventional methods such as linear/nonlinear programming, convex optimization, and algorithms based on differential calculus such as the steepest descent method should be tried. 
\item If this again fails, more established evolutionary algorithms such
as genetic algorithms and simulated annealing can be tried to tackle
the problem. 
\item If these two options do not provide any satisfactory solutions, then
try more exotic, nature-inspired metaheuristic algorithms such as
particle swarm optimization, ant-bee algorithms, or rely algorithms
or other class algorithms like Distribution Estimation, Cross-Entropy,
MCMC etc.\end{itemize}
\begin{description}
\item [{Remember\emph{.}}] \emph{It is}\textbf{\emph{ }}\emph{important
to understand why a given algorithm fails before trying a new one}. 
\end{description}
For this reason it important to get as more knowledge as possible
about the problem. Only in this way it will be possible to adapt the
best strategy to the actual problem. In this paper we will focus on
optimization strategies based on the use of Genetic Algorithms (GA).

\section{What you need to know about genetic algorithms}

Before introducing a practical discussion about optimization strategy
schemes, a few introductory comments on Genetic Algorithms are in
order. GAs are a subclass of Evolutionary Algorithms and also an example
of \emph{metaheuristic}. 
\begin{description}
\item [{Remember.}] A heuristic is a technique (consisting of a rule or
a set of rules) which seeks good solutions at a reasonable computational
cost but it does not guarantee optimality. Metaheuristic is a high-level
problem-independent algorithmic framework that provides a set of guidelines
or strategies to develop heuristic optimization algorithms. They deal
with a dynamic balance between diversification (global exploration
of different areas of the search space) and intensification (to focus
on the local fine-tune refinement of the so-far best candidate solution).
\end{description}
GAs are global, parallel, search and optimization methods that mimic
the process of natural selection developed by Charles Darwin. Evolution
is performed using a set of stochastic genetic operators, which select
individuals for reproduction, produce new individuals based on those
selected, and determine the composition of the population at the subsequent
generation. Crossover and mutation are two well known operators \citep{Fleming01geneticalgorithms}.
Terminology of GA is borrowed from natural genetic and evolution theory.
The basic idea of the evolution theory states that individuals with
a greater \textquotedblleft{}fitness to the environment\textquotedblright{}
have a greater probability of surviving and a greater probability
of winning the fights for mating. In such a way the genetic content
of the best individuals will be more and more present in the following
generations, since it will be transmitted by the offspring \citep{Leardi_2001}. 

An Individual in genetic algorithm is identified with a chromosome.
Information encoded in chromosome is called genotype. Phenotype is
values of source task variables corresponding to genotype. In other
words phenotype is decoded genotype. In the simplest codification
of a genetic algorithm chromosomes are binary string of finite length.
a \emph{gene} is a bit of this string. \emph{Allele} is value of gene,
0 or 1. One could, in principle, use any representation conceivable
for encoding the variables and indeed it strictly depends on the problem:
one needs to decide on a structure which is able to represent every
possible solution to the desired problem. When the variables are naturally
quantized, the binary GA fits nicely. However, when the variables
are continuous, it is more logical to represent them by floating-point
numbers and this is very common nowadays (for real encoding see the
chapter 3 of {[}\citep{Haupt_Haupt_2004}{]}).
\begin{description}
\item [{Remember.}] The way in which candidate solutions are encoded is
a central, if not the central, factor in the success of a genetic
algorithm. In their earlier work, Holland and his students concentrated
on binary encoding (i.e., bit strings) and GA practice has tended
to follow this lead. Much of that theory can be extended to apply
to non binary encoding, but such extensions are not as well developed
as the original theory \citep{Mitchell:1998:IGA:522098}.
\end{description}
Population is a finite set of chromosome. The fitness function (i.e.
the corresponding values of the objective function $f(x)$ in the
maximization problem or some transformations of it) compares elements
of the population, assigning each chromosome a fitness value. A chromosome
$x$ ha a better fitness with respect to $y$ if $f(x)>f(y)$. This
provides a means of ranking chromosomes from best to worst. Fitness
of individual is value of fitness function on phenotype corresponding
individual. In simple genetic algorithm size of population $N$ and
binary string length m is fixed and don\textquoteright{}t changes
in process of evolution \citep{Sharapov}. Starting with a randomly
generated population of chromosomes, a GA carries out a process of
fitness-based selection and recombination to produce a successor population,
the next generation. During recombination, parent chromosomes are
selected and their genetic material is recombined to produce child
chromosomes. After this step, in practical implementation a \emph{mutation
operator }is applied. Mutation perturbs the recombined solutions slightly
to explore their immediate neighborhood. These then pass into the
successor population. As this process is iterated, a sequence of successive
generations evolves and the average fitness of the chromosomes tends
to increase until some stopping criterion is reached. In this way,
a GA \textquotedblleft{}evolves\textquotedblright{} a best solution
to a given problem. In process of evolution one population is replaced
by another and so on, thus we select individuals with best fitness.
So in the mean each next generation (population) is fitter than it
predecessors. Genetic algorithm produces maximal fitness population,
so it solves the maximization problem (fig. 2.1).

\begin{figure}
\begin{centering}
\includegraphics[scale=0.9]{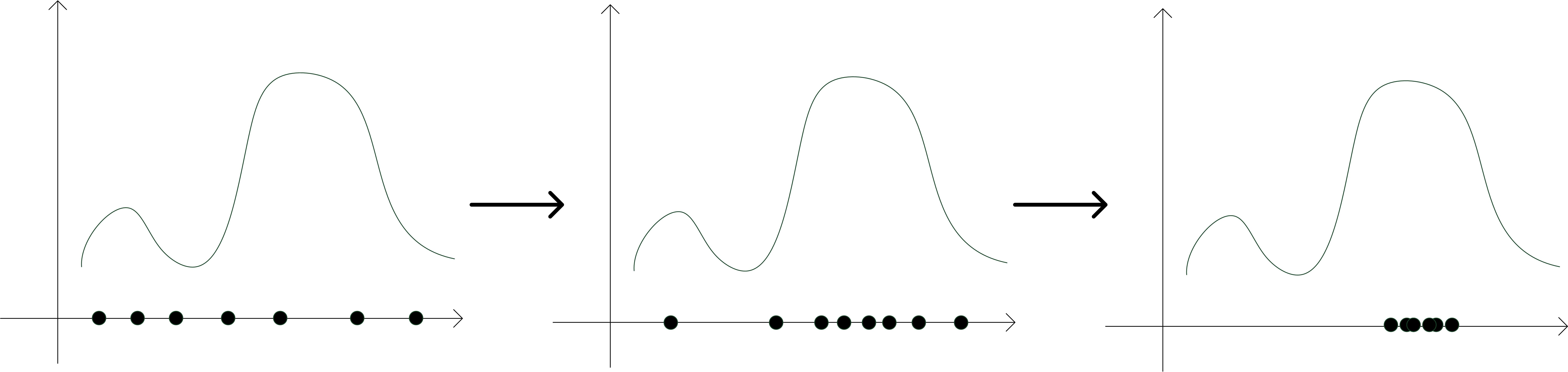}
\par\end{centering}

\caption{Generation after generation the population converges to a good solution}

\end{figure}

\begin{description}
\item [{Remember.}] Initialization is typically just creating some $N$
individuals at random. However, if something about the likely initial
\textquotedblleft{}good\textquotedblright{} regions of the search
space is known, it is a good strategy approach is to incorporate this knowledge into the initialization. In some cases can be convenient to seed the initial population with individuals selected from some sub-regions of the search domain according to the problem-specific information (\citep{Luke2009Metaheuristics}, pp. 30).
\end{description}
Summarizing, a GA works through the following steps: (1) creation
of a random initial population of $N$ potential solutions to the
problem and evaluation of these individuals in terms of their fitness
values; (2) selection of a pair of individuals as parents; (3) recombination
trough crossover of the parents, with generation of two children;
(4) replacement in the population, so as to maintain the population
number \emph{N} constant; (5) genetic mutation. Genetic operators
have been subject to intensive discussion, over both the composition
and purpose of the various operators. Essentially, crossover tends
to direct the search to superior areas of the search space, whilst
mutation acts to explore new areas of the search space and to ensure
that genetic material cannot be irretrievably lost. Choice of genetic
operators must be made together with choice of representation. Choices
of values for the parameters of the operators, such as mutation rate
and mutation size, are critical to the success of the algorithm (see
below).

Although GAs have enjoyed significant practical success, attempts
to establish a theoretical account of their precise operation have
proved more difficult. There are two goals for a satisfactory theory
of GAs. The first is to explain which classes of problem GAs are particularly
suitable for and why. The second is to provide techniques and approaches
for optimal design and implementation of GAs, as there are many choices
of structure and parameters to be made. The \emph{Schema Theorem}
\citep{Holland} has stood for a long time as a central result of
GA theory for the binary-encoded case. It attempts to explain how
the evolutionary processes in a GA can locate optimal or near-optimal
solutions, even though they sample only a tiny fraction of the set
of all possible solutions (for more details see \citep{McCall} and
\citep{Altenberg95theschema}). In essence, selection, recombination,
and mutation work together to combine small pieces of salient information
(called building blocks) from different chromosomes to form \textquotedblleft{}good\textquotedblright{}
solutions. When GA work well, chromosomes containing good building
blocks will, on average, outperform chromosomes containing inferior
building blocks. Thus, each successive generation is populated with
chromosomes containing more and more good building blocks (i.e., better
quality solutions). The goal of the GA is to ultimately find a chromosome
that is comprised of the best building blocks (i.e., an optimal solution
to the problem). When a GA converges to a non-optimal solution, it
is either because crossover does not exchange the correct material
or mutation does not explore the proper material. It has been demonstrated
that premature convergence due to mining failure is primarily due
to a poor match between encoding and genetic operators \citep{Santarelli}.

GA have demonstrated their potential in numberless projects since
they are easy to use. They are problem independent, i.e. they neither
need a special representation for candidate solutions nor put any
other restricting requirements on the problems to be tackled. Thus
they can be viewed as universally applicable. However, practice shows
that any successful application depends on on careful tuning of operators, parameters, and problem-dependent
aspects. This observation is supported by the \emph{No Free Lunch
theorems} which have shown that, on average, any deterministic or
stochastic optimization algorithm is as good as any other algorithm
on sets of problems that are closed under the permutation of fitness
values. This implies that, on the one hand, each algorithm has a niche
within the space of all problems and, on the other hand, for each
optimization problem the algorithm must be tailored to the problem
(\citep{WeickerW03}).

Practitioners must be aware that a GA is nothing else than a random search, together with a mechanism that, according to the evolutionary logic, tries to guess the next potential solution to evaluate. This can leave skeptical about the effectiveness of GAs, because there is no guarantee that such an algorithm (based on random choices) will always find the global optimum. According to \citep{Collet}, as an empirical rule in this case we can say that: 
  
\begin{itemize}
\item Given a particular maximization problem class, if the researcher already knows the best solution for a number of representative in that class and
\item if, over a significant number of runs, the proposed GA finds a solution that in average is 99\% as good as the known optimum values then, 
\item one can reasonably expect that on a new instance of the problem for which the solution is not known, the solution found by the GA  will be 99\% as good as the unknown optimum over a significant number of runs.
\end{itemize}

Despite the number of optimization techniques available nowadays the author of this paper thinks that Genetic Algorithms still play a central role in coping with a wide variety of optimization problems without any a priori assumptions about their continuity and differentiability. Even if it they still lack a strong theoretical foundation able to understand or predict the dynamics other than a superficial level, the biological analogy is conceptyally easy to understand and it is relatively simple to develop program code for any kind of application. This is common to most of nature-inspired computational techniques and can explain the speed of their diffusion virtually in every area of applied sciences far outside the field of computer sciences. Several tricks exist for improving the performance of GAs and affect the final result. Some of them will be examined in the rest of the paper.

\subsection{Implementation issues}

Several GAs have been developed; beyond the common basic idea (mimicking
the evolution of a species), they can have relevant differences. A
genetic algorithm for a particular problem must have the following
five components (\citep{CoelloThes}, pp. 94). 
\begin{enumerate}
\item A representation for potential solutions to the problem.
\item A way to create an initial population of potential solutions.
\item An evaluation function that plays the role of the environment rating
solutions in terms of their fitness. 
\item Genetic operators that alter the composition of children.
\item Values for various parameters that the genetic algorithm uses (population
size probabilities of applying genetic operators etc.)
\end{enumerate}
As already stated, representation of candidate solutions is a critical
component of a GA. One may either use the original representation
of candidate solutions or one may map the original representation
into binary strings, real-valued vectors \citep{wright:genetic} or
another representation. A good starting point is to use the original
representation for the particular problem with no significant modifications
or transformations. In the initial stage, modifications should be
done only if the chosen implementation necessitates them. If the results
obtained are not satisfactory, alternative representations should
be examined. As far as operators are concerned, different possibilities
are described in the literature. In the case of real encoding, very
common nowadays for global mono-objective optimization, an intermediate
recombination operator can be the best choice. This is usually based
on some kind of average/mixture among multiple parents: if $x=(x_{1},\ldots,x_{n})$
and $y=(y_{1},\ldots,y_{n})$ are two parents, then we can define
the component of the children:
\[
x_{i}^{\prime}=\alpha x_{i}+(1-\alpha)y_{i}\text{ }\alpha\in[0,1]
\]

The above formula can be extended to more than two parents $x_{i}^{\prime}=\alpha_{1}x_{i}+\alpha_{2}y_{i}+\alpha_{3}z_{i}+\cdots$
with $\Sigma_{i}\alpha_{i}=1$.

The mutation operator than will operate of on (a fraction of) genes
of the children. The mutation operator in the real encoded case can
be defined as component wise addition of normal distributed random
numbers. A mutation of the individual's parameters vector is calculated
as 
\[
x_{i}^{\prime}=x_{i}+\sigma N(0,1)
\]
where $N(0,1)$ is the Gaussian distribution of mean $0$ and standard
deviation $1$ and $\sigma$ is a real parameter that define the actual
standard deviation wanted%
\footnote{In some variants the value of $\sigma$ can be also adapted during
the execution of the algorithm trough a mechanism that change (usually
reduce) its value generation after generation. For more details see:
\citep{Eiben}. %
}.

There are a number of resources freely available on the Internet for
those interested in applying GAs in their own area. Tool-kits are
available in many programming languages and vary widely in the level
of programming skill required to utilize them. Those 
implementation most likely includes numerous operators and predefined set of parameters to consider; sticking with the default choices should be a good starting point.
If the results are not satisfactory, different operators may be examined
and specialized operators may be designed. There are many details that are
outside of the scope of this paper, but they can be found in specialized publications like \citep{Haupt_Haupt_2004} and \citep{Mitchell:1998:IGA:522098}. Once a GA implementation is
up and running, it is important to determine how well it is working,
and to adjust things if it is not working well. One of the simplest
analysis involves observing the top fitness vs. generation time plot.
One easy way to begin experimenting with GAs is the Genetic Algorithm
Toolbox in MATLAB \citep{Chipperfield_1995}.

\section{Typical problems in optimization }

In practice, the complexity of a given problem can be ascribed to
the following basic causes: 
\begin{itemize}
\item \emph{High number of independent variables}. The large number of candidate
solutions to an optimization problem makes it computationally very hard to be attacked by evolutionary algorithms because the number of candidate solutions grows exponentially with increasing dimensionality. This fact, which is frequently named \emph{the curse of dimensionality}, is well known by practitioners that have to handle problems with hundreds of variables. This phenomenon can be easily understood by first considering an $n$-dimensional binary search space. Here, adding another dimension to the problem means a doubling of the number of candidate solutions. So in order to obtain reliable optimization result with GA, the amount of data required to be sampled from the search space will grow exponentially with the dimensionality. The way to overcome this limitation is one of the most intriguing theoretical and practical area of research at the intersection of mathematics, statistics and computer science that has already produced a vast literature\footnote{For the non-expert in the field can be a good starting point to consult the \emph{Wikipedia} entry:\url{http://en.wikipedia.org/wiki/Curse_of_dimensionality}}.  
\item \emph{Very complex or irregular response surface}. In landscape surface
with weak (low) causality, small changes in the candidate solutions
often lead to large changes in the objective values, i. e. ruggedness.
Stated informally, a landscape is rugged if there are many local optima
of highly varying fitness concentrated in any constrained region of
the space. It then becomes harder to decide which region of the problem
space to explore and the optimizer cannot and reliable gradient information
to follow. A small modification of a very bad candidate solution may
then lead to a new local optimum and the best candidate solution currently
known may be surrounded by points that are inferior to all other tested
individuals.
\item \emph{Fitness evaluation}. Evaluating a solution in the objective
space can be by far the most computationally expensive step of any
optimization process for difficult or large-size optimization problems.
Finding the optimal solution to complex high dimensional, multi-modal
problems often requires very expensive fitness function evaluations.
For most evolutionary algorithms, a large number of fitness evaluations
(performance calculations) are needed before a well acceptable solution
can be found \citep{Jin}. In this case, it may be necessary to forgo
an exact evaluation and use an approximated fitness (also called \emph{meta-model}
or \emph{surrogate fitness}) that is computationally more efficient.
Functional surrogate models are in practice algebraic representations
of the true problem functions. The most popular ones are polynomials
(often known as \emph{response surface methodology }see\emph{ }section
4.1), Interpolation and regression polynomial techniques can be classified
in this category.Other several models are now commonly used for fitness
approximation. like the Kriging model, neural networks, including
multi-layer perceptrons, radial-basis-function networks and the support
vector machines. One can say that functional models are typically
based on the following components: a class of basis functions, a procedure
for sampling the true functions, a regression or fitting criterion,
and some deterministic or stochastic mathematical technique to combine
them all.For a comprehensive review see \citep{Fang} and \citep{Conn}.
\end{itemize}

These are very difficult cases to be attacked by evolutionary algorithms,
no special tricks exist which can directly mitigate for example the effects of rugged fitness landscapes. In GA, using large population sizes and
applying methods to increase the diversity can decrease the influence
of ruggedness, but only up to a certain degree. The lower the causality
of an optimization problem, the more rugged its fitness landscape
is, which leads to a degradation of the performance of the optimizer.
This does not necessarily mean that it is impossible to find good
solutions, but it may take very long to do so \citep{WZCN2009WIOD}.

\section{Tips for an optimization scheme}

One important objective of a preliminary optimization process is how
to get a better understanding of the objective space. A useful concept
commonly adopted in the field of metaheuristics is the notion of a
\emph{fitness landscape}, i.e. the (hyper)surface obtained by applying
the fitness function to every point in the search space over which
search is being executed. Given a specific landscape structure \textendash{}
defined by a search space, objective function, and search operators
(crossover, mutation), a GA metaheuristic can be viewed as a strategy
for navigating this structure using the information provided by the
guiding fitness function. This is a commonly used metaphor; it allows
interpreting the search in terms of well-known topographical objects
such as peaks, valleys, mesas, etc., of great utility to visualize
the behavior of the search \citep{Cotta_fromgenes}. Knowledge of
the fitness landscape structure is key to developing effective algorithms,
and consequently it has been a primary focus in the theoretical analysis
of metaheuristic methods. Here we define \emph{fitness landscape analysis}
as the step of exploring information about the structure of the problem
trough an analysis of the fitness function to be maximized. For a
continuous global optimization problem, there are several traditional
ways to categorize the objective function $f(x)$ according to some
properties like continuity, geometry, symmetries, multi-modality,
ruggedness, etc. The knowledge of these properties can give insights
about the best algorithm to use or appropriately refining the region
of interest to be sampled in the search space. For example can be
important to understand if the function is decomposable. Decomposability
is sometimes also referred to as separability, i.e functions expressed
as the sum of element functions on which small subsets of variables
have disjoint effects The optimization process of a decomposable objective
function can be performed in a sequence of \emph{N} independent optimization
processes, where each parameter is optimized independently. In the
case where the function manifest an additive decomposability property
it will be probable the search along the coordinate axes. Since optimization
methods in real life industrial design problems strictly depend on
the {}``geometry\textquotedbl{} of the problem to be solved, the
more about the response surface is known the better is possible to
tune the optimization strategy. To this end a series of tests in which
changes in the input variables are set in order to recognize
the reasons for changes in the output response.

\subsection{Fitness landscape analysis}

Response Surface Methodology (RSA) stems from the area of experimental design where the aim is to extract the maximum amount of information from a given system (represented by a mathematical function) with few and selective experiments. More technically, RSM is a collection
of mathematical and statistical techniques for empirical model building.
Trough a series of \emph{computer experiment}s carefully designed
it is possible to to identify the behavior of changes in the fitness
output. Due to the lack of data and the \textquotedbl{}curse\textquotedbl{} of the high dimensionality of the search
space, can be very difficult to obtain a perfect global functional
approximation of the original fitness function. A desirable design
of experiments should provide a distribution of points throughout
the region of interest, which means to provide as much information
as possible on the problem. The \textquotedbl{}space-filling\textquotedbl{}
methods like \emph{Latin hypercube} sampling are now easily accessible
in dedicated scientific software like MATLAB. If there is no space
for a systematic exercise of RSA, GAs can be applied preliminary as
a tool to scan the fitness landscape. A great advantage of GAs (and
other population-based strategies) over the classical techniques is
that at the end of the elaboration the user is given not just an {}``optimal\textquotedbl{}
solution, but also a population of extremely good solutions, usually
having very similar responses. 

When the optimization problem includes less than three variables, graphical methods can be fruitfully used to gain understanding on the nature of the search space. Graphical methods are often used even when the number of variables exceeds two. In that case, a practical stratagem can be to allow two variable to vary and to freeze the others. The 3D plot of the outcome can give some insights about the fitness landscape but we must be aware this approach is limited and can conduct to misleading conclusions sometimes.

\subsection{Managing the parameters}

Evolutionary algorithms are characterized by many parameters which
may be used for tuning an algorithm for a specific problem. Having
selected an encoding, there are many other choices to make. These
include: the form of the fitness function; population size; crossover
and mutation operators and their respective rates; the evolutionary
scheme to be applied; and appropriate stopping/restart conditions.
The usual design approach is a combination of experience, problem-specific
modeling and experimentation with different evolution schemes and
other parameters. Those parameters cannot be understood in isolation
but rather as an interweave network. Changing one parameter has significant
impact on the effect of other parameters. Good parameter settings
differ from problem to problem and cannot be transferred to algorithms
using operators with different characteristics. Using GA, good suggestions
for the setting of the parameters are as follows. 
\begin{description}
\item [{Population.}] An important feature of a population is its genetic
diversity: if the population is too small, the scarcity of genetic
diversity may result in a population dominated by almost equal chromosomes
and then, after decoding the genes and evaluating the objective function,
in the quick convergence of the latter towards an optimum which may
well be a local one. At the other extreme, in too large populations,
the overabundance of genetic diversity can lead to clustering of individuals
around different local optima. The population size should increase
with problem size $n$ (the number of bits or variables), and the
larger the population size, the better the obtained solution should
be \citep{Busacca_2001}. Computational efficiency issues must be
considered. The total CPU time used in an optimization run is proportional
to: (population number)$\times$(number of generation)$\times$(time
required for each fitness function evaluation). This leads to a trade-off
between having large, diverse populations that explore parameter space
widely, and having smaller populations that explore longer. In practice,
the choice is problem dependent.As a starting point for initial runs,
the population size $N_{pop}$ may be set to the problem size, $N_{pop}=n$.
If using $N_{pop}=n$ is infeasible, one may start with smaller values
of $N_{pop}$, such as $N_{pop}=50$. Changing the population size
and rerunning the GA with larger and smaller populations (e.g. $N_{pop}/2$
and $2N_{pop}$) provides an indication of whether the current population
size is adequate. If increasing the population size leads to better
solutions, the population size should be increased. If increasing
the population size leads to solutions of about the same quality,
the current population size is likely to suffice. Of course, if the
evaluation of the objective function is computationally intensive,
large populations may become intractable and, to run a sufficient
number of generations, one is often forced to use a relatively small
population size.
\item [{Reproduction.}] It is important to adopt {}``elitism'': the $k$
best individuals of each generation go directly to the next generation.
The use of elitism guarantees that the maximum fitness of the population
never decreases from one generation to the next and it normally produces
a faster convergence of the population. Furthermore, the time required
for a generation is smaller. With high elitism the risk is that all
the chromosomes are quite similar, around a good maximum, and that
it will be impossible to get out of that region: the only possibility
would be landing by chance on a higher peak, with a higher response.
Summarizing, a reproduction without elitism has a higher exploration,
while the higher the elitism the higher the exploitation. Here too,
the problem is finding a good balance. 
\item [{Crossover.}] The probability of crossover in GAs is typically quite
large. Some authors suggest that $pc=0.6$ regardless of the number
of bits $n$. Many researchers use an even larger probability of crossover,
for example, $pc=0.9$ or $pc=1$.
\item [{Mutation.}] The probability of mutation is typically set so that
the expected number of modified bits is fixed regardless the number
of parameters considered and that only a few bits are expected to
change on average. For example, one may use $pm=1/n$. Some researchers
use larger values of $pm$, but in many cases the reasons for increasing
the probability of mutation are that the remaining parameter settings
are inadequate \citep{Pelikan10geneticalgorithms}.
\end{description}

\subsection{Hybrid strategies}

Experimental results show that for most GAs (initialized with random
values), evolution makes extremely rapid progress at first, as the
diverse elements in the initial population are combined and tested.
Over time, the population begins to converge, with the separate individuals
resembling each other more and more. Effectively this results in the
GA narrowing its search in the objective space and reducing the size
of any changes made by evolution until eventually the population converges
to a single solution. In contrast, {}``hill-climbing\textquotedbl{}
or {}``gradient ascent\textquotedbl{} algorithms, which initialize
at some point and then move greedily toward the nearest local optima,
are maximally exploitative, since they make no attempt to explore
alternative solutions away from its current trajectory. In hybrid
algorithms, a GA with good exploration capacities are often used to
locate some promising zones within the wide search space, while the
local optimization methods exploit the located promising zones to
achieve the best solution quickly and accurately. So the hybrid algorithms
not only have the exploration capabilities, parallelism and combination
of GAs, but also obtain the exploitation power of local search methods
\citep{hibri}. Also Holland himself, the {}``father\textquotedbl{}
of genetic algorithms, suggested that the genetic algorithm should
be used as a pre-processor for performing the initial search, before
invoking a local search method to optimize the final population. Simply
stated, the theory suggests that the key role of the global searcher
is to find one or a number of good regions and the key role of the
local searcher is to and local optima within these regions. Typically
a GA followed by quasi deterministic gradient search type algorithm
is the easiest choice when the fitness is known to be differentiable.
If not, other possibilities for a local search step can be explored
from the family of random search like the\emph{ Simulated Annealing}
algorithm%
\footnote{In this case, parameters of SA must be tuned carefully to avoid expensive
global search overlap.%
} (see appendix 1). So summarizing we have: 

\noindent \begin{center}
GeneticAlgorithm+LocalSearchAlgorithm=HybridAlgorithm.
\par\end{center}

One of the key issues in the design of efficient hybrid procedures
is the appropriate coordination of the global and local search. In
this case it is important how to manage the improved solution that
is produced by the local search.The two main possibilities are captured
in the literature by the terms \emph{Lamarckian }and \emph{Baldwinian
hybrid} \citep{Krasnogor2006}. In a Lamarckian hybrid, the local
search procedure again uses the global procedure\textquoteright{}s
value as a starting point, searches until reaching an optimum, and
then passes back the function value found with back-substitution of
the solution value found. That is, suppose that individual $x$ belongs
to the population $P_{t}$ in generation $t$ and that the fitness
of $x$ is $f(x)$. Furthermore, suppose that the local search produces
a new individual $x^{\prime}$ with a better fitness ($f(x^{\prime})>f(x)$
for the maximization problem). The designer of the algorithm will
replaces $x$ with $x^{\prime}$, in which case $P_{t+1}=P_{t}\backslash\left\{ x\right\} \cup\left\{ x^{\prime}\right\} $and
the genetic information in $x$ is lost and replaced with that of
$x^{\prime}$. In a Baldwinian hybrid, the local search procedure
uses the GA value as a starting point, searches until reaching an
optimum, and then passes back the function value found without back-substituting
the solution value found. In this case the genetic information of
$x$ is kept but its fitness altered: $f(x)=f(x^{\prime})$. At the
first sight, Lamarckian hybrids may seem more intuitive, but in some
cases back-substitution of solution values found by local search has
been found to reduce solution variance to a point such that subsequent
exploration is diminished. Various rules of thumb have suggested Baldwinian
moves with a small percentage of Lamarckian moves as desirable in
practice (for a comprehensive analysis on this see \cite{hibri}). Finding an adequate way to split resources between the global
and local search requires a systems-level understanding of the roles
of the two procedures.

\section{Appendix 1}

\subsubsection*{Simulated Annealing%
\footnote{See chapter 12 of \citep{yang}%
}}

The simulated annealing algorithm for optimization imitates the annealing
process in metallurgy, a technique involving heating and controlled
cooling of a material to increase the size of its crystals and reduce
defect. By analogy with this physical process, each step of the simulated
annealing algorithm replaces the current solution by a new solution
with a probability that depends on the difference of objective function
values at the two solution points and a parameter called the temperature.
A simulated annealing algorithm always accepts moves that decrease
the value of the objective function (to be maximized). Moves that
increase the value of the objective function are accepted with probability
$p=e^{\Delta/T}$ where $\Delta$ is the change in the value of the
objective function and $T$ is the temperature. A random number generator
that generates numbers distributed uniformly on the interval $(0,1)$
is sampled, and if the sample is less than p, the move is accepted.
The temperature $T$ is initially high. Therefore, the probability
of accepting a move that increases the objective function is initially
high. The temperature is gradually decreased as the search progresses.
I.e., the system is cooled slowly. In the end, the probability of
accepting a move that increases the objective function becomes vanishingly
small. In general, the temperature is lowered in accordance with an
annealing schedule procedure like for example $T_{k+1}=\lyxmathsym{\textgreek{b}}T_{k},\text{ }0<\lyxmathsym{\textgreek{b}}<1$. 

\bibliographystyle{hc-en}
\bibliography{bib}

\end{document}